\documentclass{article}

\usepackage{arxiv}
\usepackage[utf8]{inputenc} % allow utf-8 input
\usepackage[T1]{fontenc}    % use 8-bit T1 fonts
\usepackage{hyperref}       % hyperlinks
\usepackage{url}            % simple URL typesetting
\usepackage{booktabs}       % professional-quality tables
\usepackage{amsfonts}       % blackboard math symbols
\usepackage{microtype}      % microtypography
\usepackage{graphicx}       % allow images inclusion
\usepackage{float}          % allow img positioning 
\usepackage{xcolor}
\usepackage{amsmath}        % 
\usepackage{mathtools}      %
\usepackage{tcolorbox}
\tcbuselibrary{theorems}

\newtcbtheorem{mytheo}{Sifrian at a Glance - Version}
{colback=white!5,colframe=gray!35!black,fonttitle=\bfseries}{th}
\usepackage{graphicx,wrapfig,lipsum}
\title{Exact Stochastic Second Order Deep Learning
}

\author{
 Fares B.~Mehouachi\thanks{Dr. Fares B. Mehouachi is Lead Machine Learning Researcher at the Directed Energy Research Centre, Technology Innovation Institute, Abu Dhabi, United Arab Emirates, email: fares.mehouachi@tii.ae.} \\
 Directed Energy Research Centre\\
 Technology Innovation Institute\\
 United Arab Emirates\\
   \And
 Chaouki Kasmi\thanks{Dr. Chaouki Kasmi is Chief Researcher and Researcher in Statistical Electromagnetics at the Directed Energy Research Centre, Technology Innovation Institute, Abu Dhabi, United Arab Emirates,  email: chaouki.kasmi@tii.ae.}\\
 Directed Energy Research Centre\\
 Technology Innovation Institute\\
 United Arab Emirates \\
}

\begin{document}
\maketitle

\begin{abstract}
Optimization in Deep Learning is mainly dominated by first-order methods which are built around the central concept of backpropagation. Second-order optimization methods, which take into account the second-order derivatives are far less used despite superior theoretical properties. This inadequacy of second-order methods stems from its exorbitant computational cost, poor performance, and the ineluctable non-convex nature of Deep Learning. Several attempts were made to resolve the inadequacy of second-order optimization without reaching a cost-effective solution, much less an exact solution. In this work, we show that this long-standing problem in Deep Learning could be solved in the stochastic case, given a suitable regularization of the neural network. Interestingly, we provide an expression of the stochastic Hessian and its exact eigenvalues. We provide a closed-form formula for the exact stochastic second-order Newton direction, we solve the non-convexity issue and adjust our exact solution to favor flat minima through regularization and spectral adjustment. We test our exact stochastic second-order method on popular datasets and reveal its adequacy for Deep Learning.
\end{abstract}

% keywords can be removed

\section*{Key Contributions}
To ease the reading of this paper, we provide here the key contributions of this work.
\begin{itemize}
     \item The Sifrian: a second-order Lagrangian for Deep Learning;
    \item Exact solution for the stochastic Newton direction for Feed-Forward Networks;
    \item Adjustment of the exact stochastic Newton direction for non-convex optimization; 
    \item Explicit expression for the stochastic Hessian;
    \item Exact eigenvalues of the stochastic Hessian ;
    \item Spectral adjustment of the Hessian through regularization for targeting flat minima;
    \item Testing the exact stochastic second-order Deep Learning on MNIST and F-MNIST data sets; 
\end{itemize}
\newpage

\section{Introduction}
Deep Learning has gained tremendous momentum in the last two decades thanks to its ability to outperform other Machine Learning techniques in discovering hidden patterns in high dimensional data and learning abstract representation with minimal human intervention~\cite{lecun2015deep}. Deep learning, at its core, requires the adjustment of several weight matrices and bias vectors governing the learning process between several interconnected layers: for a given input such as an image file or an audio recording, the data is transformed into a vector and then successively modified, at each layer, by a weight matrix multiplication and a bias vector addition. The transition from one layer to another is finalized by the application of an activation function, mimicking the biological behavior of brain neurons. The weights and biases of the different layers of the neural network are chosen as a solution to an optimization problem that reduces the mismatches between the outputs of the last layer and the labels associated with the input data. Finding optimal parameters of the network (i.e. weights and biases) for a given dataset is an NP-hard problem~\cite{judd1988complexity}. In practice, the hardness is circumvented using iterative methods such as gradient descent methods which generate weights and biases updates that reduce output errors incrementally.

The cornerstone of Deep Learning is the backpropagation algorithm which allows for efficient computation of the errors' gradient in linear time with respect to the network parameters size. Backpropagation could be derived as a solution to a constrained optimization using Lagrange multipliers. Several adjoint-state vectors (multipliers) are introduced in an augmented cost function that describes the mechanics of the neural network. The adjoint-state vectors are selected in a way that simplifies the computation of the gradient. This is achieved if the adjoints verify a specific system of equations similar to the main system of equations of the underlying neural network~\cite{rumelhart1986learning,lecun1988theoretical}. The adjoint state equations have the particularity of being solvable backward from the output layer to the initial layers hence the appellation backpropagation. The backpropagation algorithm is fast and yields an exact value of the gradient without implementing any finite difference schemes or computing superfluous Fréchet derivatives.

Unfortunately, backpropagation remains essentially a first-order method and tends to exhibit severe limitations once the depth (i.e. the number of layers) of the neural networks increases. Most notably, vanishing and exploding gradients are two prevalent problems of any first-order method. Furthermore, the convergence of gradient descent (i.e. 1st order methods) is notoriously slow on very deep networks resulting generally in under-fitting and a poor performance/accuracy. Several improvements have been devised to accelerate the convergence of the first-order gradient descent such as the momentum method~\cite{nesterov1983method}, Adam~\cite{kingma2014adam} or adaptive regularization~\cite{duchi2011adaptive}. Currently, the state-of-the-art optimizer for very deep neural networks is probably a combination of the stochastic gradient descent and the momentum method~\cite{sutskever2013importance}.

Devising efficient learning algorithms is fundamental in Artificial Intelligence and this led naturally to the exploration of second-order optimization as a potential candidate that might reach superior performances. The premise of second-order methods is that correcting the gradient direction using curvature information (i.e. second-order derivatives), might yield quadratic convergence, and might significantly accelerate learning beyond what any first-order method could reach. Such a statement is valid only when the underlying optimization problem is convex. Unfortunately, Deep Learning is a non-convex problem, which is a major hurdle that conceptually limits the direct adoption of second-order methods. In practice, second-order methods for Deep Learning are known to be inadequate, due to the high computational cost and the poor performance~\cite{lecun2012efficient}. Despite these limitations, several approaches and variations have been developed to circumvent such shortcomings. The existing second-order methods could be classically divided into two categories: the approximate Newton methods and the Hessian-free methods. The approximate second-order methods a.k.a quasi/truncated Newton methods are based on building a proxy estimation of the inverted Hessian matrix which exhibits desirable features such as positivity and invertibility. Quasi-Newton~\cite{bordes2009sgd}, iterative (BFGS) algorithm~\cite{bollapragada2018progressive}, or Hessian diagonal estimation are examples of this type of approximate methods. The main drawback of these previously mentioned methods is their approximate nature and the associated computational cost which remains relatively high compared to simple gradient descent methods. The other type of second-order method is called Hessian free and it is based on adjoint-state techniques which allow computing the product of the Hessian with any vector~\cite{pearlmutter1994fast,martens2010deep}. This method applies a differential operator called the $\mathcal{R}$-operator to the backpropagation equations~\cite{pearlmutter1994fast} and allows the computation of the Hessian product with any given vector at the additional cost of one forward and one backward passes through the neural network. The $\mathcal{R}$-operator is dependent on the pre-selected vector, to which Hessian multiplication is applied. In this context, the inversion step remains problematic since finding the adequate Newton direction which reproduces the gradient’s direction after multiplication by the Hessian is not trivial.  Finding the Newton direction is usually carried through the conjugate gradient method which is typically stopped after sampling few orthogonal directions instead of covering the full space~\cite{martens2010deep}. This approach requires a few dozens of forward and backward passes for the inversion which slows down the learning process. A different approach for approximating the curvature information (Fisher matrix) has been introduced more recently through Kronecker Factorization (K-FAC)~\cite{martens2015optimizing}, this method is efficient because it approximates the curvature in a way that is straightforward to invert. K-FAC could be probably considered as the current state-of-the-art second-order method. Despite the numerous attempts to bridge the knowledge gap, it is reasonable to consider that all existing second-order methods are approximations.

\begin{figure}[H]
    \centering
    \includegraphics[width=50mm]{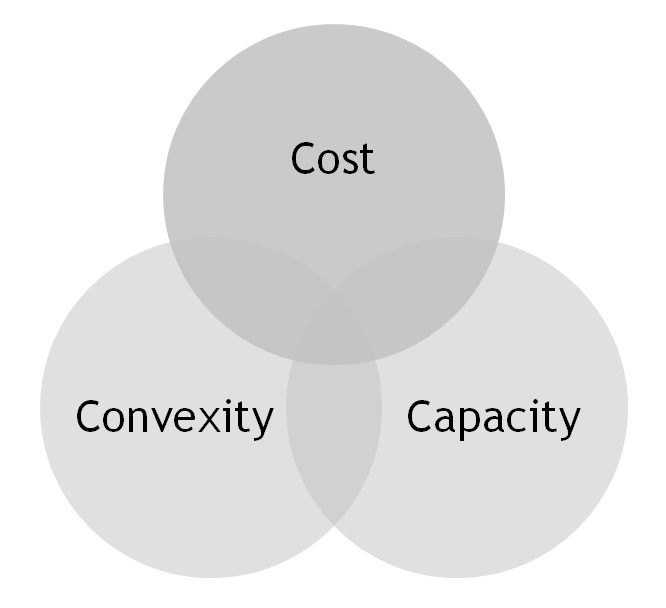}
    \caption{The 3C of second-order Deep Learning.}
    \label{fig:FFN_3c}
\end{figure}

In this paper we focus on solving the inherent inadequacy of second order method for Deep Learning. We define three criteria that we think would characterize adequacy for second order Deep learning. These criteria are the following three "C"s: Cost, Convexity and Capacity (3C diagram shown in Fig.~\ref{fig:FFN_3c}):
\begin{itemize}
\item \underline{\textbf{C}ost} efficiency, linear if possible;
\item \underline{\textbf{C}onvexity} indifference;
\item \underline{\textbf{C}apacity} to generalize beyond the training set.
\end{itemize}

To satisfy the three Cs of second-order Deep Learning, we revisit, in this work, the concept of backpropagation, and develop a similar framework that characterizes the Newton direction, through a second-order Lagrangian which we name the Sifrian. \\

The Sifrian requires four additional adjoint state variables and is built in a way that brings forth the Hessian without explicitly computing it. The system generated by the Sifrian could be used to reproduce the effect of the $\mathcal{R}$-operator. We solve the equations of the Sifrian in the stochastic case and provide an exact closed-form solution to the Newton direction under regularization constraints of the state variable. The closed-form nature of our solution guarantees Cost efficiency. We also tackle the problem of non convexity and we adjust our solution to ensure that it is a descent direction. We provide also an explicit formula for the Hessian and we further adjust our regularization to guarantee that the stochastic Hessian radius is minimal, which is necessary to target the flat minima. This extra adjustment is a desirable feature to enhance the Capacity of generalization. We finally test our exact stochastic method on popular datasets to prove its adequacy for second-order Deep Learning.

The paper is organized as follows: in Section~\ref{sec:FNNN}, the formalism of Feed-Forward Neural Network is introduced as a standard model for Deep Learning and we derive the backpropagation method to compute the gradient of FNN using the Lagrangian. In Section~\ref{sec:Sifrian}, we propose a new Lagrangian for second-order optimization, named the Sifrian, which will allow the characterization of the Newton update for Deep Learning using four additional adjoint-state vectors. The differential properties of the Sifrian are explored and a closed-form solution for the Newton direction in the stochastic case is presented in Section~\ref{sec:NewtonDirect}. We further readjust the exact solution to make it suitable for non-convex optimization and we additionally provide a dumped version for extra stability. In Section~\ref{sec:Hadjust}, we derive an expression of the Hessian and focus on minimizing its radius to favor flat minima. In Section~\ref{sec:Testcases}, two test cases are proposed to demonstrate the applicability of our method for second-order Deep Learning. Conclusions are finally drawn in Section~\ref{sec:conclusion} along with open research opportunities.

\section{Feed-Forward Neural Network and Backpropagation}
\label{sec:FNNN}
In this section, we recall the main details of the Feed-Forward Neural Network (FNN) which will be used as a standard model for Deep Learning. Other networks such as the Convolutional Neural Networks (CNN) or Recurrent Neural Networks (RNN) could be derived as special cases of the FNN. The notation used throughout this paper is similar to notations presented in~\cite{lecun1988theoretical}.

\subsection{The Feed-Forward Neural Network (FNN) model}
The main equation governing the FNN, a.k.a. the forward model is the following:
\begin{equation}
x_{p}(k)=F\left(W(k)\,x_{p}(k-1)+\beta(k)\right),\quad k=1..n, \quad p\in D    
\end{equation}
where $D$ is the database, $p$ designs one single pattern from the database (e.g. one single image or audio recording), $k$ is the layer number and $n$ is the total number of layers in the network. The initial input for pattern $p$ is $x_{p}(0)$ (e.g. the vectorized input image data). The state variable $x_{p}$ is transformed at each layer $k$ through a multiplication by a weight matrix $W(k)$ and an addition of a bias vector $\beta(k)$. The weight and bias, defined for each layer, are the parameters of the neural network, and do not depend on the analysed pattern $p$. 

\begin{figure}[h]
    \centering
    \includegraphics[width=80mm]{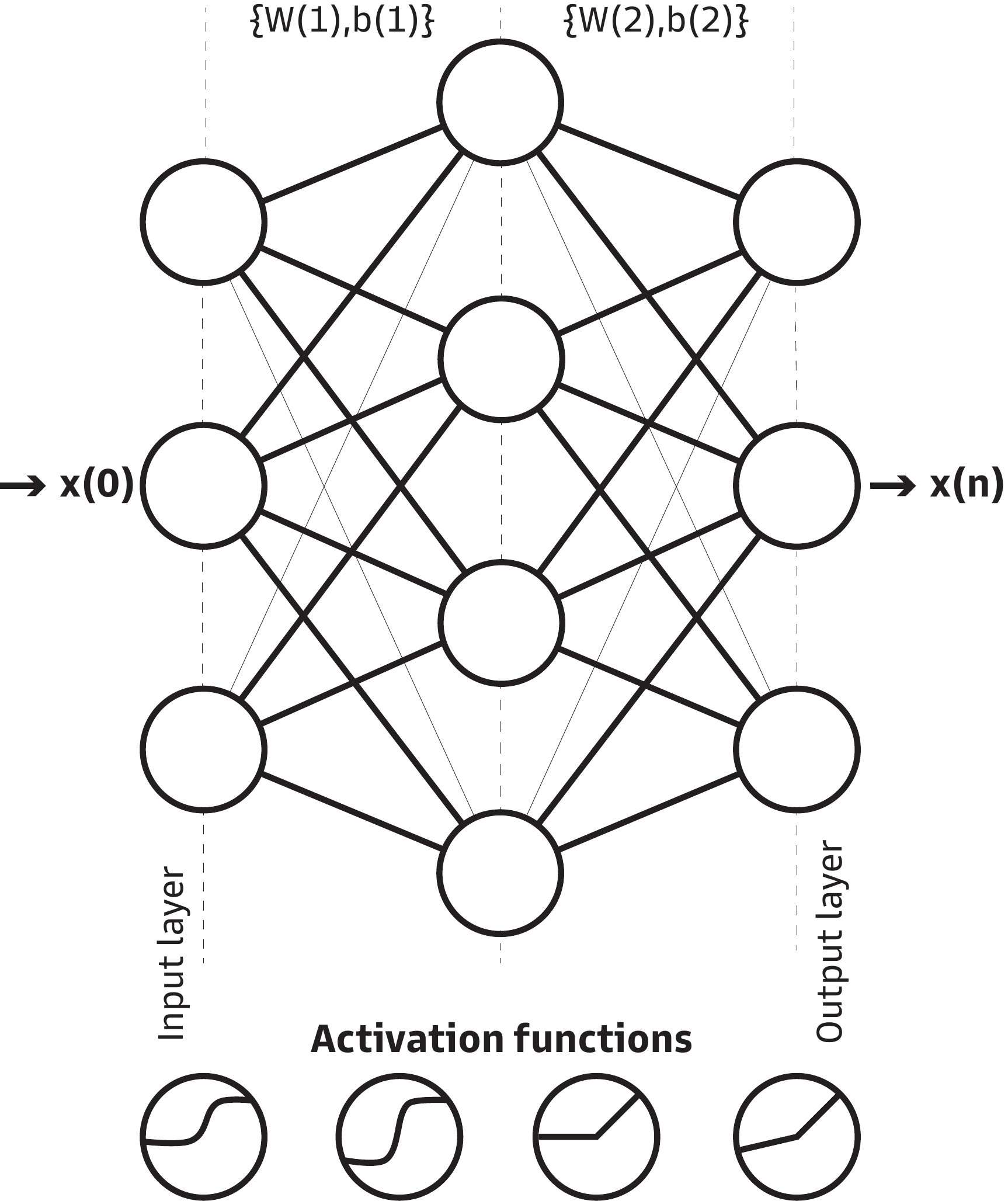}
    \caption{Diagram representation of a Feed Forward Network (FNN) with one hidden layer. The listed activation functions are from left to right: sigmoid, tanh, ReLU and leaky ReLU.}
    \label{fig:FFN_sketch}
\end{figure}

The activation function $F$, which is typically a sigmoid or a Rectified Linear Unit (ReLU) depicted in Fig.~\ref{fig:FFN_sketch}, is applied element wise on the resulting vector $a_{p}(k)=W(k)\,x_{p}(k-1)+\beta(k)$ and serves to introduce non linearity in the neural network.

The final output of propagation of the pattern $p$ through the neural network $x_{p}(n)$ is compared with the desired label $d_p$, and a cost (loss) function $J$ is formulated from the mismatches between the outputs and the labels of the entire database. Different types of cost function exist in the literature, we present below the least-squares one for the sake of simplicity:
\begin{equation}
    J\left(x(W,\beta)\right)=\sum_{p\in D}\frac{1}{2}\left(d_{p}-x_{p}(n)\right)^{T}\left(d_{p}-x_{p}(n)\right)
\end{equation}
The parameters of the network $\left\{ W,\beta\right\} _{k=1..n}$ are updated progressively to minimize the cost function. The sensitivity of the cost function $J$ to small variations of the parameters $\left\{ W,\beta\right\} _{k=1..n}$ is the gradient $\left\{ \frac{dJ}{dW(k)},\frac{dJ}{d\beta(k)}\right\}_{k=1..n} $. 

Given that the biases are vectors and the weights are matrices, and in order to get coherent notations we vectorize the weight matrix $\left\{W(k)\right\}_{k=1..n}$. As such the forward model becomes:
\begin{equation}
    x_{p}(k)=F\left(W(k)\,x_{p}(k-1)+\beta(k)\right)=F\left(\left(x_{p}(k-1)^{T}\otimes I_{k-1}\right)\mathsf{Vec}\left[W(k)\right]\,+\beta(k)\right)
\end{equation}

The $\text{Vec}$ operator aligns the columns of the matrix $W(k)$ into one single column, the $\otimes$ is the Kronecker product, and $I_{k}$ is the square matrix with the same dimension as the $k^{th}$ layer. 

Both variations of the forward model should be useful for computing partial derivatives, and we will pick the most suitable form depending on the desired outcome dimension. Within this context the network parameter we want to optimize is a vector concatenation of the vectorized weights and biases:
\begin{equation}
    \begin{array}{c}
w=\left[\begin{array}{ccccc}
\text{Vec}\left[W(1)\right]^{T} & \cdots & \text{Vec}\left[W(k)\right]^{T} & \cdots & \text{Vec}\left[W(n)\right]^{T}\end{array}\right]^{T}\\
\\
\beta=\left[\begin{array}{ccccc}
\beta(1)^{T} & \cdots & \beta(k)^{T} & \cdots & \beta(n)^{T}\end{array}\right]^{T}
\end{array}
\end{equation}

It is typical to merge the weight and bias parameters in only one optimization vector to get compact formulas. However, we intentionally keep the two variables separate, despite the extra level of complexity. This particular choice is crucial for building the exact solution in the next sections.

Before proceeding further with optimization for Deep Learning, we recall the two following properties of the Kronecker product which will be useful in Section~\ref{sec:Hadjust}. The first property elucidates the difference between the vectorization of a matrix $A\in\mathbb{R}^{m,n}$ and the vectorization of its transpose $A^T$.

Let $U_{m,n}$ be the orthogonal perfect shuffle permutation Matrix ~\cite{SteebandHardy}, then we have the following property:
\begin{equation}
\text{Vec}(A)~=~U_{m,n}\text{Vec}(A^{T})
\end{equation}
It is worth noting that if $(m=1)$ or $(n=1)$ then $U_{m,n}$ simplifies to the identity matrix.

The second useful property of the Kronecker product is related to the inversion of the product terms order. For two matrices $A_{1}$ in $\mathbb{R}^{{m_1},{n_1}}$ and $A_{2}$ in $\mathbb{R}^{{m_2},{n_2}}$, the permutation formula is given by:
\begin{equation}
A_{2}~\otimes~A_{1}~=~U^{T}_{{m_1},{m_2}}(A_{1}~\otimes~A_{2})U_{{n_1},{n_2}}
\end{equation} 

These two properties are important to carry on calculations in Section~\ref{sec:Hadjust} and to go beyond the gradient descent. 

The choice of the gradient direction as a minimisation direction stems for the first order approximation of the cost function. A small perturbation $\left\{ \delta w,\delta \beta\right\}$ around central network parameters $\left\{ w_0,\beta_0\right\}$ yields the following approximation:
\begin{equation}
J\left(w_{0}+\delta w,\beta_{0}+\delta\beta\right)=J\left(w_{0},\beta_{0}\right)+\left\langle \delta w,\frac{dJ}{dw}\right\rangle +\left\langle \delta\beta,\frac{dJ}{d\beta}\right\rangle +\mathcal{O}\left(\left\Vert \delta w\right\Vert _{2}^{2},\left\Vert \delta\beta\right\Vert _{2}^{2}\right)
\end{equation}

The $\left\langle .\,,.\right\rangle$ operator is the canonical real scalar product which also could be expressed using matrix scalar product. In this case the weight perturbation term could be expressed as follows:
\begin{equation}
    \left\langle \delta w,\frac{dJ}{dw}\right\rangle =\sum_{k=1..n}\left\langle \delta w(k),\frac{dJ}{dw(k)}\right\rangle =\sum_{k=1..n}\mathsf{trace}\left(\delta W(k)\frac{dJ}{dW(k)}^{T}\right)
\end{equation}

The $\mathsf{trace}$ is the matrix trace operator. With these notation, as long as the gradient $\left\{ \frac{dJ}{dw},\frac{dJ}{d\beta}\right\} $ is not null, updating the network parameters along the opposite direction of the gradient is guaranteed to decrease the cost function at least locally. Computing the gradient through direct differentiation of the cost function $J$ with respect to (w.r.t.) the parameters is impractical as it requires the knowledge of the Fréchet derivatives which is the sensitivity matrix (or tensor) of the state vector w.r.t. the different weights or biases. 

\begin{equation}
    \frac{dJ}{dW(k)}=-\sum_{p}\left(\frac{dx_{p}(n)}{dW(k)}\right)^{T}\left(d_{p}-x_{p}(n)\right)
\end{equation}

The quantity $\left(\frac{dx_{p}(n)}{dW(k)}\right)_{k=1..n}$ is not trivial to compute, changes with each pattern $p$ and is expensive to store. Deep learning would be impractical if not for the backpropagation algorithm which we derive in the next section. 

\subsection{The Lagrangian and backpropagation for Deep Learning: }
\label{sec:regul}
The origin of backpropagation could be traced back to the early 1970s and could be derived by casting Deep Learning as a constrained optimization problem. This section is similar to~\cite{lecun1988theoretical}, except that we add a regularization term dependent on the state variable $\mathrm{R}(x)$ to the cost function $J$:
 \begin{equation}
     J_{R}=J+\mathrm{R}\left(x\right)
 \end{equation}
 
This additional term has to verify the following invertibility condition:
 \begin{equation}
  \left(\frac{\partial J_R}{\partial x(m)\partial x(k)}\right)_{k,m=1..n} \quad \text{is\, definite}
  \label{eq:1}
 \end{equation}

Such a choice might seem unusual since the regularization often concerns the network parameters such as the weights and the biases. We explicitly avoid any regularization on the network parameters $\left\{ W,\beta\right\} _{k=1..n}$ as this is crucial to derive simople solution for second-order optimization in Section~\ref{sec:NewtonDirect}.
  
 We call a regularization term that verifies Equ.~\ref{eq:1} admissible. An example of such a regularization is the following function:
 \begin{equation}
     \mathrm{R}(x)=\frac{\lambda}{2}\sum_{p\in D}\sum_{k=1..n-1}\left\langle x_{p}(k),x_{p}(k)\right\rangle 
 \end{equation}

The previous term is admissible as long as the constant $\lambda$ is non-null. This regularization term could be changed at will and any choice is acceptable as long as it verifies the admissibility condition defined by Equ.~\ref{eq:1}. 

In order to derive the backpropagation algorithm we introduce the following Lagrangian as per the notation of~\cite{lecun1988theoretical}:
 \begin{equation}
\begin{array}{c}
\mathcal{L}(x,W,\beta,b)=\sum_{p\in D}\frac{1}{2}\left\Vert d_{p}-x_{p}(n)\right\Vert _{2}^{2}+\sum_{p\in D}\sum_{k=1..n}\left\langle x_{p}(k)-F\left(W(k)\,x_{p}(k-1)+\beta(k)\right),b_{p}(k)\right\rangle \\
\\
+\frac{\lambda}{2}\sum_{p\in D}\sum_{k=1..n-1}\left\Vert x_{p}(k)\right\Vert _{2}^{2}
\end{array}
 \end{equation}

The Lagrangian contains the original least square cost function, the admissible regularization term, and the product of the forward equation with the adjoint state vectors $\left(b_{p}(k)\right)_{p\in D,\,k=1..n}$. These adjoint vectors are defined for each layer of the network and each pattern of the database, however, their values are not fixed yet and will be chosen later in a way that simplifies the gradient computation. If the state variable $x$ verifies the forward equation then the Lagrangian simplifies to the cost function:
\begin{equation}
    \mathcal{L}(x(W,\beta),W,\beta,b)=J_{R}(x(W,\beta))
\end{equation}

The total derivative of the previous Lagrangian w.r.t. the weights or biases is the gradient and could be expressed in terms of partial derivatives as follows:
\begin{equation}
\begin{array}{c}
\frac{d\mathcal{L}(x(W,\beta),W,\beta,b)}{dW(k)}=\frac{\partial\mathcal{L}}{\partial W(k)}+\sum_{p\in D}\sum_{m=1..n}\left(\frac{dx_{p}(m)}{dW(k)}\right)\left(\frac{\partial\mathcal{L}}{\partial x_{p}(m)}\right)=\frac{dJ_{R}}{dW(k)}\\
\\
\frac{d\mathcal{L}(x(W,\beta),W,\beta,b)}{d\beta(k)}=\frac{\partial\mathcal{L}}{\partial\beta(k)}+\sum_{p\in D}\sum_{m=1..n}\left(\frac{dx_{p}(m)}{d\beta(k)}\right)\left(\frac{\partial\mathcal{L}}{\partial x_{p}(m)}\right)=\frac{dJ_{R}}{d\beta(k)}
\end{array}
\end{equation}

The partial derivatives of the Lagrangian w.r.t to $\left\{ W,\beta\right\} _{k=1..n}$ are straightforward to compute. The Fréchet derivatives $\left(\frac{dx_{p}(m)}{dW(k)},\frac{dx_{p}(m)}{d\beta(k)}\right)_{p\in D,\,k,m=1..n}$ are not trivial to evaluate and the core idea of backpropagation is the selection of the adjoint state variables $\left(b_{p}(k)\right)_{p\in D,\,k=1..n}$ which cancels the superfluous terms. Such a simplification is achievable if:
\begin{equation}
    \left(\frac{\partial\mathcal{L}}{\partial x_{p}(k)}\right)_{p\in D,\,k=1..n}=0
\end{equation}

Computing the previous partial derivative yields the following backpropagation equation:
\begin{equation}
    \frac{\partial \mathcal{L}}{\partial x_{p}(k)}=\left(x_{p}(n)-d_{p}\right)\mathbf{1}_{k=n}+\mathbf{1}_{k=1..n-1}\lambda x_{p}(k)+b_{p}(k)-\mathbf{1}_{k=1..n-1}W(k+1)^{T}\nabla F\left(a_{p}(k+1)\right)b_{p}(k+1)=0
\end{equation}
$\mathbf{1}_E$ is the indicator function and it is equal to one if the underlying condition $E$ is true and null otherwise. The nabla operator $\nabla F$ is a diagonal square matrix with element wise derivative of its argument (along the diagonal). We added a new variable $\left(a_{p}(k)\right)_{p\in D,k=1..n}$ to further simplify the notations which corresponds to the input of the selected activation function.
\begin{equation}
    a_{p}(k)=W(k)x_{p}(k-1)+\beta(k)
\end{equation}

The resolution of the backpropagation system could be split into a boundary condition and a backward propagation system as follows:
\begin{equation}
    \begin{cases}
b_{p}(n)=\left(d_{p}-x_{p}(n)\right) & \text{for}\; k=n\\
\\
b_{p}(k-1)=W(k)^{T}\nabla F\left(a_{p}(k)\right)b_{p}(k)-\lambda x_{p}(k-1) & \text{for}\; k=1..n-1
\end{cases}
\end{equation}

This particular choice of the adjoint state vectors $\left(b_{p}(k)\right)_{p\in D,\,k=1..n}$ yields a simple formula for the gradient of the cost function:
\begin{equation}
\begin{cases}
G(k)=\frac{\partial\mathcal{L}}{\partial W(k)}=-\sum_{p}\nabla\left[F\left(a_{p}(k)\right)\right]b_{p}(k)x_{p}(k-1)^{T}\\
\\
g(k)=\frac{\partial\mathcal{L}}{\partial\beta(k)}=-\sum_{p}\nabla F\left(a_{p}(k)\right)b_{p}(k)
\end{cases}
\end{equation}

The backpropagation algorithm is the core engine of Deep Learning, it is simple, elegant, and provides a straightforward solution to an otherwise tedious problem to solve. Unfortunately, developing a second-order backpropagation method remained elusive, and the current state-of-the-art is an augmented backpropagation that allows computing the product of the Hessian with a given vector, without explicitly calculating or storing the Hessian. In the following section, we provide a different framework for backpropagation which allows the characterization of the second-order Newton direction.

\section{Beyond the First Order Lagrangian: the Sifrian}
\label{sec:Sifrian} 
The main novelty in this work starts from this section onward. The characterization of the Hessian product with a vector has been developed for Deep Learning using the $\mathcal{R}$-operator. We provide a new approach to the equation governing the second-order update using a different formalism. To this end, we introduce a similar function to the Lagrangian which we will refer to as Sifrian.
 
\subsection{The Sifrian}
It is legitimate to wonder if the magic of the Lagrangian could be replicated to a higher order. The Lagrangian combines the cost function and the forward model multiplied by some adjoint variables which we are free to choose.
A natural extension would be to create a function that has the forward model, the backward model, and also the definition of the gradient. We omit the cost function, as its derivation will generate the gradient again. Starting from this paradigm, we introduce the Sifrian\footnote{The choice of the Sifrian appellation stems from the Arabic word "Sifr", which means zero, and it is also the characteristic value of the Sifrian when all the equations of Deep Learning are satisfied. For Francophones, the syllable "rian" is changed to "rien" which means literally "nothing". Subjectively, we simply like the name!}, the equivalent Lagrangian for second-order optimization:

\begin{equation}
\begin{array}{c}
\mathcal{S}(x,b,W,\beta,G,g,\gamma,\zeta,N,\eta)=\sum_{k,p}\left\langle x_{p}(k)-F\left(a_{p}(k)\right),\gamma_{p}(k)\right\rangle \\
\\
+\sum_{k,p}\left\langle \left(x_{p}(n)-d_{p}\right)\mathbf{1}_{k=n}+\mathbf{1}_{k=1..n-1}\lambda x_{p}(k)+b_{p}(k)-\mathbf{1}_{k=1..n-1}W(k+1)^{T}\nabla F\left(a_{p}(k+1)\right)b_{p}(k+1),\zeta_{p}(k)\right\rangle \\
\\
+\sum_{k}\left\langle G(k)+\sum_{p}\nabla\left[F\left(\,a_{p}(k)\right)\right]b_{p}(k)x_{p}(k-1)^{T},N\left(k\right)\right\rangle \\
\\
+\sum_{k}\left\langle g(k)+\sum_{p}\nabla F\left(a_{p}(k)\right)b_{p}(k),\eta\left(k\right)\right\rangle 
\end{array}
\end{equation}

The Sifrian, might seem intimidating at a first glance: it has a all the equations generated by the first order backpropagation, multiplied by four new adjoint parameters $\left\{ \gamma_{p,k},\zeta_{p,k},N_{k},\eta_{k}\right\} $. While the Lagrangian describes a saddle point, the Sifrian, describes an equilibrium and it has a unique null value when all the forward, backward and gradients equations are verified i.e.:
\begin{equation}
\mathcal{S}(x\left(W,\beta\right),W,\beta,b\left(W,\beta,x\right),G\left(W,\beta,x,b\right),\gamma,\zeta,N,\eta)=0
\end{equation}

We call the previous quantity the equilibrated Sifrian and it is critical to notice its invariance to total derivation w.r.t. either the weights or biases. Expressing the total derivative of the equilibrated Sifrian using partial derivatives yields the following equations:
\begin{equation}
\begin{cases}
\frac{d}{dW(k)}\mathcal{S}=\frac{\partial\mathcal{S}}{\partial W(k)}+\sum_{m,p}\frac{dx_{p}(m)}{dW(k)}\frac{\partial\mathcal{S}}{\partial x_{p}(m)}+\frac{db_{p}(m)}{dW(k)}\frac{\partial\mathcal{S}}{\partial b_{p}(m)}+\sum_{m}\frac{dG(m)}{dW(k)}\frac{\partial\mathcal{S}}{\partial G(m)}+\frac{dg(m)}{dW(k)}\frac{\partial\mathcal{S}}{\partial g(m)}=0\\
\\
\frac{d}{d\beta(k)}\mathcal{S}=\frac{\partial\mathcal{S}}{\partial\beta(k)}+\sum_{m,p}\frac{dx_{p}(m)}{d\beta(k)}\frac{\partial\mathcal{S}}{\partial x_{p}(m)}+\frac{db_{p}(m)}{d\beta(k)}\frac{\partial\mathcal{S}}{\partial b_{p}(m)}+\sum_{m}\frac{dG(m)}{d\beta(k)}\frac{\partial\mathcal{S}}{\partial G(m)}+\frac{dg(m)}{d\beta(k)}\frac{\partial\mathcal{S}}{\partial g(m)}=0
\end{cases}
\label{eq:zpartial}
\end{equation}

As with the first order backpropagation, we have the new adjoints variables $\left\{ \gamma_{p,k},\zeta_{p,k},N_{k},\eta_{k}\right\}_{p\in D,\,k=1..n} $ which we can select freely, to simplify the superfluous terms. We need concretely to use the adjoints to cancel the effect of the Fréchet derivatives. This in turn requires that the new adjoints satisfy the following equations:
\begin{equation}
\begin{cases}
\frac{\partial\mathcal{S}}{\partial x_{p}(k)}=0 & p\in D,\,k=1..n\\
\\
\frac{\partial\mathcal{S}}{\partial b_{p}(k)}=0 & p\in D,\,k=1..n
\end{cases}
\end{equation}

In this case, Equ.~\ref{eq:zpartial} could be cast as the following block matrix system:
\begin{equation}
\left[\begin{array}{c}
\frac{\partial\mathcal{S}}{\partial W(k)}\\
\frac{\partial\mathcal{S}}{\partial\beta(k)}
\end{array}\right]+\left[\begin{array}{cc}
\frac{dG}{dW} & \frac{dg}{dW}\\
\frac{dG}{d\beta} & \frac{dg}{d\beta}
\end{array}\right]\left[\begin{array}{c}
\frac{\partial\mathcal{S}}{\partial G}\\
\frac{\partial\mathcal{S}}{\partial g}
\end{array}\right]=0
\end{equation}

The matrix that appears in the previous system is nothing else but the Hessian. The derivative of the Sifrian w.r.t. to the gradient is straightforward to compute and involves the two adjoints $\left\{ N,\eta\right\} $. 

The Characteristic system of the Sifrian becomes:
\begin{equation}
\left[\begin{array}{c}
\frac{\partial\mathcal{S}}{\partial W(k)}\\
\frac{\partial\mathcal{S}}{\partial\beta(k)}
\end{array}\right]+\left[\begin{array}{cc}
\frac{dJ_{R}}{dWdW} & \frac{dJ}{dWd\beta}\\
\frac{dJ_{R}}{d\beta dW} & \frac{dJ}{d\beta d\beta}
\end{array}\right]\left[\begin{array}{c}
N\\
\eta
\end{array}\right]=0
\end{equation}

If we constrain the new adjoints to verify the two following equation:
\begin{equation}
\begin{cases}
\frac{\partial\mathcal{S}}{\partial W(k)}=-G(k) & k=1..n\\
\\
\frac{\partial\mathcal{S}}{\partial\beta(k)}=-g(k) & k=1..n
\end{cases}
\end{equation}

Then, $\left\{ N,\eta\right\} $ are bound to be the Newton direction we are searching for:
\begin{equation}
H\left[\begin{array}{c}
N\\
\eta
\end{array}\right]=\left[\begin{array}{c}
G\\
g
\end{array}\right]
\end{equation}

In sum, the introduction of the Sifrian leads to a system of four equations, which we need to solve to compute the Newton direction.

\subsection{The Sifrian equations}
The four equations of the Sifrian which characterize the Newton direction are the following:
\begin{equation}
\begin{cases}
\frac{\partial\mathcal{S}}{\partial x_{p}(k)}=0 & p\in D,\,k=1..n\\
\\
\frac{\partial\mathcal{S}}{\partial b_{p}(k)}=0 & p\in D,\,k=1..n\\
\\
\frac{\partial\mathcal{S}}{\partial W(k)}=-G(k) & k=1..n\\
\\
\frac{\partial\mathcal{S}}{\partial\beta(k)}=-g(k) & k=1..n
\end{cases}
\end{equation}

We carry the derivation and we report hereafter the Sifrian tetrad system:
\begin{equation}
\begin{cases}
\sum_{p}-\nabla F\left(a_{p}(k)\right)\gamma_{p}\left(k\right)x_{p}(k-1)^{T}-\mathbf{1}_{k=2..n}\nabla F\left(a_{p}(k)\right)b_{p}(k)\zeta_{p}(k-1)^{T}\\
+\sum_{p}\nabla^{2}\left[F\left(a_{p}(k)\right)\right]b_{p}(k)\left[N\left(k\right)x_{p}(k-1)+\eta(k)-\mathbf{1}_{k=2..n}W(k)\zeta_{p}(k-1)\right]x_{p}(k-1)^{T}\\
=\sum_{p}\nabla\left[F\left(a_{p}(k)\right)\right]b_{p}(k)x_{p}(k-1)^{T}\\
\\
\sum_{p}-\nabla F\left(a_{p}(k)\right)\gamma_{p}\left(k\right)+\sum_{p}\nabla^{2}\left[F\left(a_{p}(k)\right)\right]b_{p}(k)\left[N\left(k\right)x_{p}(k-1)+\eta\left(k\right)-\mathbf{1}_{k=2..n}W(k)\zeta_{p}(k-1)\right]\\
=\sum_{p}\nabla F\left(a_{p}(k)\right)b_{p}(k)\\
\\
\gamma_{p}\left(k\right)-\mathbf{1}_{k=1..n-1}W(k+1)^{T}\nabla F\left(a_{p}(k+1)\right)\gamma_{p}\left(k+1\right)+\mathbf{1}_{k=n}\zeta_{p}(n)+\lambda\mathbf{1}_{k=1..n-1}\zeta_{p}(k)\\
+\mathbf{1}_{k=1..n-1}W(k+1)^{T}\nabla^{2}\left[F\left(a_{p}(k+1)\right)\right]b_{p}(k+1)\left[N\left(k+1\right)x_{p}(k)+\eta(k+1)-\mathbf{1}_{k=1..n-1}W(k+1)\zeta_{p}(k)\right]\\
+\mathbf{1}_{k=1..n-1}N\left(k+1\right)^{T}\nabla\left[F\left(a_{p}(k+1)\right)\right]b_{p}(k+1)=0\\
\\
\zeta_{p}(k)-\mathbf{1}_{k=2..n}\nabla F\left(a_{p}(k)\right)W(k)\zeta_{p}(k-1)+\nabla\left[F\left(a_{p}(k)\right)\right]\left[N\left(k\right)x_{p}(k-1)+\eta\left(k\right)\right]=0
\end{cases}
\end{equation}

The operator $\nabla^2 F$ is a third-order super-diagonal tensor, which yields a matrix after multiplication with a vector. The Sifrian tetrad system involves several layers, several patterns, and might seem difficult to solve. Despite this impression, simple closed formulas for the previous system exist. Before moving further into the resolution, it is important to notice the similarities with the outcomes of the $\mathcal{R}$-Operator.

\subsection{Comparison with the $\mathcal{R}$-operator} 
Characterizing the effect of the Hessian in Deep Learning without explicitly computing its values nor storing them, has been proposed by~\cite{pearlmutter1994fast}, using an abstract operator $\mathcal{R}$ which requires the preselection of a specific direction to study its transformation through the Hessian. The Sifrian tetrad is capable of achieving the same goal without introducing any extra layers of abstraction. In fact if we relax the two gradient constraints from the Sifrian tetrad, and we preselect the values of $\left\{ N,\eta\right\} $, then we get the following system:
\begin{equation}
\ensuremath{H\left[\begin{array}{c}
N\\
\eta
\end{array}\right]=\left[\begin{array}{c}
-\frac{\partial\mathcal{S}}{\partial W(k)}\\
-\frac{\partial\mathcal{S}}{\partial\beta(k)}
\end{array}\right]}
\end{equation}

In practice using the values of $\left\{ N,\eta\right\} $, the last Sifrian tetrad equation allows computing the value of $\zeta$ through a forward pass. Once the values of $\zeta $ are known, a backward pass using the third tetrad equation allows us to get the values of $\gamma$. The product of the Hessian with the preselected direction $\left\{ N,\eta\right\} $ is finally computed using the first two equations of the tetrad.
\begin{equation*}
\left\{ N,\eta\right\} \rightarrow\zeta\rightarrow\gamma\rightarrow H\left\{ N,\eta\right\} 
\end{equation*}

Aside from its similarity with the Lagrangian, the main added value of the Sifrian formulation over the $\mathcal{R}$-operator is the added clarity and visibility about the role of each variable. This extra level of understanding is what will allow us to find the exact solution in the following section.

\section{Closed Form Formulas for the Stochastic Newton Direction}
\label{sec:NewtonDirect} 
Knowledge of closed-form solutions, when possible, is of the utmost importance to any physical or mathematical problem. Most often closed-form solutions do not exist or are unknown. Few PDEs (Partial Derivatives Equations) have a known closed-form solution such as the heat equation, Black and Scholes EDP in finance, wave equation in a homogeneous media. A closed-form solution delivers supreme computational performance and allows further understanding of the sensitivity of the solution to any of its parameters. A closed-form solution for second-order Deep Learning is unheard of to the best of our knowledge. We will provide such a solution in the stochastic case. We also handle the null space which appears naturally since the Hessian is non-invertible.	

\subsection{Case of the piecewise linear activation function}
The Sifrian tetrad system might seem complicated yet it significantly simplifies once the activation function is piecewise linear, i.e. its second derivative is null almost everywhere such as the ReLu function. We  further consider the stochastic optimization problem where only one pattern $p$ is optimized at each step. In this scenario the Sifrian system simplifies considerably:
\begin{equation}
\begin{cases}
\begin{array}{c}
-\nabla F\left(a_{p}(k)\right)\gamma_{p}\left(k\right)x_{p}(k-1)^{T}-\mathbf{1}_{k=2..n}\nabla F\left(a_{p}(k)\right)b_{p}(k)\zeta_{p}(k-1)^{T}=\nabla\left[F\left(a_{p}(k)\right)\right]b_{p}(k)x_{p}(k-1)^{T}\end{array}\\
\\
\begin{array}{c}
-\nabla F\left(a_{p}(k)\right)\gamma_{p}\left(k\right)=\nabla F\left(a_{p}(k)\right)b_{p}(k)\end{array}\\
\\
\begin{array}{c}
\gamma_{p}\left(k\right)-\mathbf{1}_{k=1..n-1}W(k+1)^{T}\nabla F\left(a_{p}(k+1)\right)\gamma_{p}\left(k+1\right)+\mathbf{1}_{k=n}\zeta_{p}(n)+\lambda\mathbf{1}_{k=1..n-1}\zeta_{p}(k)\end{array}\\
+\mathbf{1}_{k=1..n-1}N\left(k+1\right)^{T}\nabla\left[F\left(a_{p}(k+1)\right)\right]b_{p}(k+1)=0\\
\\
\begin{array}{c}
\zeta_{p}(k)-\mathbf{1}_{k=2..n}\nabla F\left(a_{p}(k)\right)W(k)\zeta_{p}(k-1)+\nabla\left[F\left(a_{p}(k)\right)\right]\left[N\left(k\right)x_{p}(k-1)+\eta_{p}\left(k\right)\right]=0\end{array}
\end{cases}
\end{equation}

\begin{figure}[h]
    \centering
    \includegraphics[width=140mm]{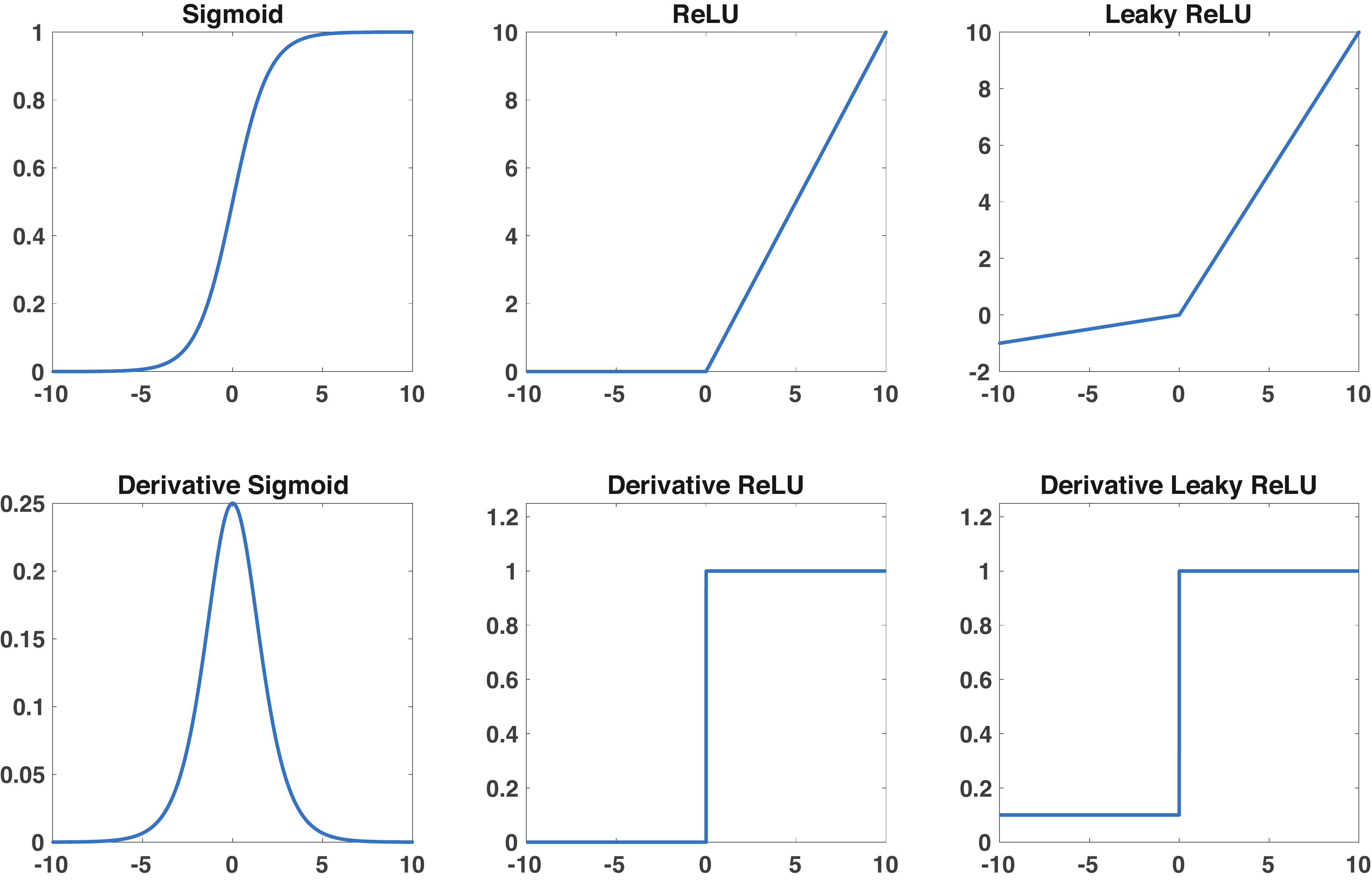}
    \caption{Examples of activation functions and their derivatives.}
    \label{fig:activations}
\end{figure}

Most importantly, the second equation establishes a direct relationship between the first-order adjoint $b$ and one of the second-order adjoints $\gamma$. This second equation, derived from the bias perturbation, will trigger a cascade of simplifications in the Sifrian system. We explicitly chose in Section~\ref{sec:FNNN} to distinguish the handling of the weights from the biases for this exact reason. The first necessary condition that emerges at this stage is the invertibility of $\nabla F$, and this can be achieved by selecting a strictly monotonous activation function such as the leaky ReLU function instead of a classic ReLU (represented Fig.~\ref{fig:activations}). 

If we assume the strict monotony of the activation function then the first two equations simplify to:
\begin{equation}
\begin{array}{c}
\gamma_{p}\left(k\right)=-b_{p}\left(k\right)\\
\\
\mathbf{1}_{k=1..n-1}b_{p}(k+1)\zeta_{p}(k)^{T}=0
\end{array}
\end{equation}

The rank one matrix above is null if and only if $b_{p}$ or $\zeta_p$ is null. The $b_{p}$ is the backpropagated errors and it can be null if for e.g. the weights are initialized at zero. We consider the case where the adjoint $b_{p}(k)$ is null as degenerate. If the adjoint vectors are non null then, we get the following simplified system:
\begin{equation}
\begin{cases}
\begin{array}{c}
\mathbf{1}_{k=1..n-1}\zeta_{p}(k)=0\end{array}\\
\\
\begin{array}{c}
\gamma_{p}\left(k\right)=-b_{p}(k)\end{array}\\
\\
\begin{array}{c}
\mathbf{1}_{k=1..n-1}\lambda x_{p}(k)+\mathbf{1}_{k=1..n-1}N\left(k+1\right)^{T}\nabla\left[F\left(a_{p}(k+1)\right)\right]b_{p}(k+1)=0\end{array}\\
\\
\begin{array}{c}
\mathbf{1}_{k=n}\left(d_{p}-x_{p}(n)\right)-\nabla\left[F\left(a_{p}(k)\right)\right]\left[N\left(k\right)x_{p}(k-1)+\eta_{p}\left(k\right)\right]=0\end{array}
\end{cases}
\end{equation}

The Newton weight is a matrix and it is determined through its product with the adjoint vector: so it is only characterized through one direction, hence determining the Newton direction is a degenerate problem but the minimal rank solution is still unique, and it is the following:
\begin{equation}
\begin{cases}
N\left(k\right)=-\lambda\frac{\nabla F\left(a_{p}(k)\right)b_{p}(k)x_{p}(k-1)^{T}}{\left\Vert \nabla F\left(a_{p}(k)\right)b_{p}(k)\right\Vert _{2}^{2}}\quad k=2..n\\
\\
\eta(k)=\nabla\left[F\left(a_{p}(k)\right)\right]^{-1}\left(d_{p}-x_{p}(n)\right)\mathbf{1}_{k=n}-N\left(k\right)x_{p}(k-1)
\end{cases}
\label{eq:sol}
\end{equation}

\begin{figure}[h]
    \centering
    \includegraphics[width=75mm]{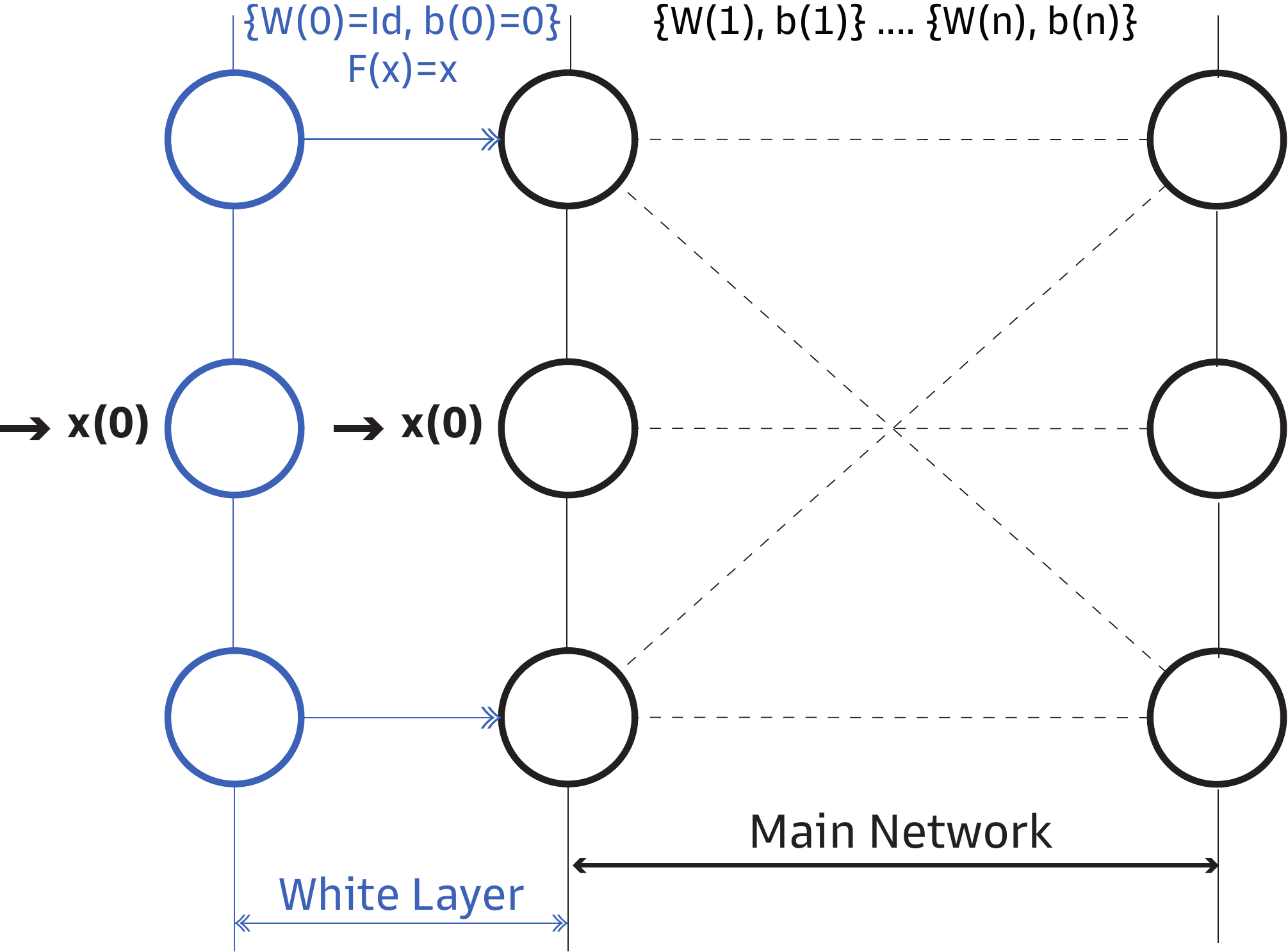}
    \caption{White layer extension of the feed forward neural network.}
    \label{fig:white_layer}
\end{figure}

The null space is of large dimension and it is a hyperspace. It is also crucial to notice that the characterization of Newton weight $N(k)$ starts only from the second layer. One way to circumvent such a difficulty is to add an extra white layer, depicted Fig.~\ref{fig:white_layer}, which consists in multiplication by the identity matrix. The white layer is included in the optimization as the first (input) layer. In this case, the solution for the Newton weight could be extended to the first layer of the main network too.

To summarize, if we assume that:
\begin{itemize}
\item The cost function is regularized;
\item The activation function is strictly monotonous;
\item The first adjoint vectors are non-null;
\end{itemize}
then, a unique rank one solution for the second-order stochastic optimization exists and it is straightforward to compute. Even in the degenerate case, where at least one of the three conditions is not satisfied, the previous solution is still valid, but elements from the null space could be added to it without restrictions.

The introduction of the regularization is crucial to get a non-null weight update for the different layers. If the regularization is removed (i.e. $\lambda=0$) a very granular solution is given by our exact solution: only the bias of the last layer is updated to compensate for the mismatch between the label and network output.

Despite the similarity, the Newton direction and the gradient are different because the adjoint term $b_p(k)$ is not the same. This difference comes from the regularization parameter $\lambda$ which plays a similar role to the step length in the closed-form solution of the Newton direction. 

The results of this section extends also to any strictly monotonous activation function (i.e. non piecewise linear) case. In fact, the terms involving $\nabla^2$ do not hinder the simplification of the second-order adjoints. $\mathbf{1}_{k=1..n-1}\zeta_{p}(k)$ remain null, and this yields similar formula to the piecewise linear case except for the last layer bias. At this stage we were able to provide a simple closed-form solution which scales the gradient, however, it is important to notice that it is still not suitable for Deep Learning due to the non-convexity issue. We adapt our solution for the non-convex optimization in the next section.

\subsection{Escaping the saddles: a descent direction}
The previous solution is simple, cost-effective and has a minimal rank but it is not necessarily a descent direction. To illustrate this we provide a different (but equivalent) characterization of the stochastic Newton direction. Starting from the Sifrian system, and after simplification we obtain:
\begin{equation}
\begin{cases}
\begin{array}{c}
\lambda x_{p}(k-1)+N\left(k\right)^{T}\nabla\left[F\left(a_{p}(k)\right)\right]b_{p}(k)=0\end{array}\\
\\
\begin{array}{c}
\mathbf{1}_{k=n}\left(d_{p}-x_{p}(n)\right)+\nabla\left[F\left(a_{p}(k)\right)\right]\left[N\left(k\right)x_{p}(k-1)+\eta_{p}\left(k\right)\right]=0\end{array}
\end{cases}
\end{equation}

It is important to notice that fragments of the gradient are present within the characterization of the Newton direction. Multiplied by the bias gradient, we get the following characterization:
\begin{equation}
\begin{cases}
\begin{array}{c}
\lambda G(k)-\nabla\left[F\left(a_{p}(k)\right)\right]b_{p}(k)b_{p}(k)^{T}\nabla\left[F\left(a_{p}(k)\right)\right]N\left(k\right)=0\end{array}\\
\\
\begin{array}{c}
\begin{array}{c}
g(k)g(k)^{T}\eta_{p}\left(k\right)=\left(-\lambda\left\Vert x_{p}(k-1)\right\Vert _{2}^{2}+\mathbf{1}_{k=n}\left\Vert d_{p}-x_{p}(n)\right\Vert _{2}^{2}\right)g(k)\end{array}\end{array}
\end{cases}
\end{equation}

Rearranging the previous terms, we get the following characterization:
\begin{equation}
\begin{cases}
\begin{array}{c}
G(k)=\frac{g(k)g(k)^{T}}{\lambda}N\left(k\right)\end{array}\\
\\
\begin{array}{c}
g\left(k\right)=\frac{g(k)g(k)^{T}}{\left(-\lambda\left\Vert x_{p}(k-1)\right\Vert _{2}^{2}+\mathbf{1}_{k=n}\left\Vert d_{p}-x_{p}(n)\right\Vert _{2}^{2}\right)}\eta(k)\end{array}
\end{cases}
\label{eq:adhoc}
\end{equation}

The sign of $\lambda$ is important and it dictates if the Newton direction is descendant or ascendant. From the previous equation, it is clear that the Newton direction cannot be a descent direction both in weight and bias. If we choose $\lambda$ positive to minimize the norm of the state vector $\left(x_p(k)\right)_{k=1..n}$, then the bias of the Newton direction is not suitable for Deep Learning. 

A simple and efficient solution consists in inverting the signs of non-descent directions to avoid saddle points \cite{dauphin2014identifying}. With this approach we get the following characterization:
\begin{equation}
\begin{cases}
\begin{array}{c}
G(k)=\frac{g(k)g(k)^{T}}{\lambda}N_{MK}\left(k\right)\end{array}\\
\\
\begin{array}{c}
g\left(k\right)=\frac{g(k)g(k)^{T}}{\left(\lambda\left\Vert x_{p}(k-1)\right\Vert _{2}^{2}+\mathbf{1}_{k=n}\left\Vert d_{p}-x_{p}(n)\right\Vert _{2}^{2}\right)}\eta_{MK}(k)\end{array}
\end{cases}
\end{equation}
We added the subscript MK (short for Mehouachi and Kasmi) to differentiate our solution from the Newton direction. The rank one MK solution is unique and has the following formula:

\begin{equation}
\begin{cases}
\begin{array}{c}
N_{MK}\left(k\right)=\lambda\frac{g(k)x_{p}(k-1)^{T}}{g(k)^{T}g(k)}\end{array}\\
\\
\begin{array}{c}
\eta_{MK}(k)=\frac{\left(\lambda\left\Vert x_{p}(k-1)\right\Vert _{2}^{2}+\mathbf{1}_{k=n}\left\Vert d_{p}-x_{p}(n)\right\Vert _{2}^{2}\right)g(k)}{g(k)^{T}g(k)}\end{array}
\end{cases}
\end{equation}
The MK solution is guaranteed to be a descent direction and hence would escape saddle points. Similar to the Newton solution it allows for the proper scaling of the gradient both in weight and bias. This is possible thanks to the introduction of the regularisation term $\lambda$, which remains however to be selected.

\subsection{The damped solution}

We introduced in the previous section a perturbation to the Newton direction to escape concave directions that would hinder the minimization. The division by norm of the backpropagated errors could be problematic if the errors are small. To avoid an exploding update, we add the following damping term to the equation (\ref{eq:adhoc}):
\begin{equation}
\begin{cases}
\begin{array}{c}
G(k)=\left(\mu I+\frac{g(k)g(k)^{T}}{\lambda}\right)N\left(k\right)\end{array}\\
\\
\begin{array}{c}
g\left(k\right)=\left(\mu I+\frac{g(k)g(k)^{T}}{\left(\lambda\left\Vert x_{p}(k-1)\right\Vert _{2}^{2}+\mathbf{1}_{k=n}\left\Vert d_{p}-x_{p}(n)\right\Vert _{2}^{2}\right)}\right)\eta_{MK}(k)\end{array}
\end{cases}
\end{equation}

The resolution of the previous system is carried through the Shermann-Morisson inversion formula, which yields the following expression:
\begin{equation}
\begin{cases}
N_{MK}(k)=\frac{\lambda}{\mu\lambda+\left\Vert \nabla F\left(a_{p}(k)\right)b_{p}(k)\right\Vert _{2}^{2}}G(k)\\
\\
\eta_{MK}(k)=\frac{\left\{ \mathbf{1}_{k=n}\left\Vert d_{p}-x_{p}(n)\right\Vert _{2}^{2}+\lambda\left\Vert x_{p}(k-1)\right\Vert _{2}^{2}\right\} }{\mu\left\{ \mathbf{1}_{k=n}\left\Vert d_{p}-x_{p}(n)\right\Vert _{2}^{2}+\lambda\left\Vert x_{p}(k-1)\right\Vert _{2}^{2}\right\} +\left\Vert \nabla F\left(a_{p}(k)\right)b_{p}(k)\right\Vert _{2}^{2}}g(k)
\end{cases}
\end{equation}

The main use of damping is to avoid numerical division by zero. Large values of damping $\mu$ would naturally favor a direction similar to the gradient. Null damping would yield back the MK direction. 

\section{The Hessian and its Spectral Adjustment}
\label{sec:Hadjust} 

\subsection{The Hessian of stochastic Deep Learning}
We provided, so far in this work, an exact expression of the Newton direction and a saddle free version which we named the MK direction. Looking further into the Sifrian tetrad system of equations, it is possible to derive an explicit expression of the Hessian in the stochastic case. Moreover, we reshape the Hessian into an LDL by block decomposition, which gives us unprecedented direct access to the spectrum of the Stochastic Deep Learning Hessian. This spectrum is closely related to the regularization we selected. Leveraging our knowledge about the spectrum, we choose a regularization that minimizes the Hessian radius which will favor flat minima~\cite{hochreiter1995simplifying,foret2020sharpness}. This approach allows us to define optimal regularization parameters, which reduce the number of hyper-parameters selection. For a better spectral adjustment we relax the regularization parameter and allow it to vary across neural layers i.e.: $\lambda\rightarrow\left(\lambda_{k}\right)_{k=1..n}$. This increase in regularization parameters is related to the following cost function:
\begin{equation}
    J=\frac{1}{2}\left\Vert d_{p}-x_{p}(n)\right\Vert _{2}^{2}+\sum_{k=1..n-1}\frac{\lambda_{k}}{2}\left\Vert x_{p}(k)\right\Vert _{2}^{2}
\end{equation}

In order to express the Hessian, we first start by assembling all the vector variable across all the layers into one single compact structure:
\begin{equation}
\left\{ \zeta_{p}(1),\zeta_{p}(2),...,\zeta_{p}(n)\right\} \rightarrow\left[\begin{array}{c}
\zeta_{p}(1)\\
\zeta_{p}(2)\\
\vdots\\
\zeta_{p}(n)
\end{array}\right]=\zeta_p
\end{equation}

The state vector $x_p(k)$ is also assembled but starts from the input $x_p(0)$. We add a tilde to emphasize the lag:
\begin{equation}
\bar{x}_{p}=\left[\begin{array}{c}
x_{p}(0)\\
\vdots\\
x_{p}(k-1)\\
\vdots\\
\vdots\\
x_{p}(n-1)
\end{array}\right]
\end{equation}
where $n$ is the number of layer. 

Let's also introduce the backward operator $\mathcal{I}-\mathcal{N}_{p}$ is:
\begin{equation}
\mathcal{I}-\mathcal{N}_{p}=\left[\begin{array}{ccccc}
I_1 & 0 & \cdots & \cdots & 0\\
-W(2)\nabla F\left(a_{p}(1)\right) & \ddots & \ddots & 0 & \vdots\\
0 & -W(k)\nabla F\left(a_{p}(k-1)\right) & I_k & \ddots & \vdots\\
\vdots & \ddots & \ddots & \ddots & 0\\
0 & \cdots & 0 & -W(k)\nabla F\left(a_{p}(n-1)\right) & I_n
\end{array}\right]
\end{equation}

The backward operator is divided into two elements: the first is the identity $\mathcal{I}$ which is the concatenation of all the identity matrix for each layer $\left\{ I_{k}\right\} _{k=1..n}$. The operator $\mathcal{N}_p$ corresponds the sub block-diagonal entries and it forms a nilpotent matrix. While we will avoid any inversion throughout this paper for the sake of efficiency, it is important to notice that the inverse of the backward operator is straightforward to compute:
\begin{equation}
    \left[I-\mathcal{N}_{p}\right]^{-1}=\sum_{i=0}^{n-1}\mathcal{N}_{p}^{i}
\end{equation}

We introduce the diagonal derivative matrix $\nabla F_p$:
\begin{equation}
\nabla F_p=\left[\begin{array}{cccccc}
\nabla F\left(a_{p}(1)\right)\\
 & \ddots\\
 &  & \nabla F\left(a_{p}(k)\right)\\
 &  &  & \ddots\\
 &  &  &  & \ddots\\
 &  &  &  &  & \nabla F\left(a_{p}(n)\right)
\end{array}\right]
\end{equation}
The regularisation matrix $\Lambda$:
\begin{equation}
\Lambda=\left(\begin{array}{ccccccc}
\lambda_{1}I_{1}\\
 & \lambda_{2}I_{2}\\
 &  & \ddots\\
 &  &  & \lambda_{k}I_{k}\\
 &  &  &  & \ddots\\
 &  &  &  &  & \lambda_{n-1}I_{n-1}\\
 &  &  &  &  &  & I_{n}
\end{array}\right)
\end{equation}

In order to get a compact representation, we introduce the following matrix: $\bar{X}=\text{diag}\left(\bar{x}_{p}\left(k\right)\otimes\nabla F_{p}\left(k\right)\right)_{k=1..n}$:
\begin{equation}
    \bar{X}=\left[\begin{array}{ccccc}
\bar{x}_{p}\left(1\right)\otimes\nabla F_{p}\left(1\right)\\
 & \ddots\\
 &  & \bar{x}_{p}\left(k\right)\otimes\nabla F_{p}\left(k\right)\\
 &  &  & \ddots\\
 &  &  &  & \bar{x}_{p}\left(n\right)\otimes\nabla F_{p}\left(k\right)
\end{array}\right]
\end{equation}
We introduce another matrix based on the first adjoints vectors $B=\text{sdiag}\left\{ \mathbf{1}_{k=2..n}I_{k-1}\otimes F\left(a_{p}(k)\right)b_{p}(k)\right\} $:
\begin{equation}
    B=\left[\begin{array}{ccccc}
0\\
I_{1}\otimes\nabla F\left(a_{p}(2)\right)b_{p}(2) & \ddots\\
 &  & 0\\
 &  &  & \ddots\\
 &  &  & I_{n-1}\otimes\nabla F\left(a_{p}(n)\right)b_{p}(n) & 0
\end{array}\right]
\end{equation}

Most of the Kronecker products are straightforward to derive except the $B$ due to the transposition operation. Few permutations matrices appear as per Section~\ref{sec:FNNN} but could be absorbed in the matrix term $B$. With the previous notations, the Sifiran tetrad system becomes the following KKT matrix: 
\begin{equation}
\left[\begin{array}{cccc}
\nabla F_{p} & 0 & 0 & 0\\
\bar{X} & B & 0 & 0\\
0 & \left\{ \mathcal{I}-\mathcal{N}_{p}\right\}  & \nabla F_{p} & \bar{X}^{T}\\
\left\{ \mathcal{I}-\mathcal{N}_{p}\right\} ^{T} & \Lambda & 0 & B^{T}
\end{array}\right]\left[\begin{array}{c}
\gamma\\
\zeta\\
\eta\\
N
\end{array}\right]=\left[\begin{array}{c}
g\\
G\\
0\\
0
\end{array}\right]
\end{equation}

The main $M$ and peripheral $P$ matrices are:
\begin{equation}
M=\left[\begin{array}{cc}
0 & \left\{ \mathcal{I}-\mathcal{N}_{p}\right\} \\
\left\{ \mathcal{I}-\mathcal{N}_{p}\right\} ^{T} & \Lambda
\end{array}\right];\quad P=\left[\begin{array}{cc}
\nabla F_{p} & 0\\
\bar{X} & B
\end{array}\right]
\end{equation}
The matrix notations we introduce in this section along with the Sifrian (KKT) matrix system are the main tools we will use to change the spectrum of the Hessian matrix.

In Section~\ref{sec:regul}, we emphasized the importance of having an admissible regularization, yet until this point, we have not explicitly used it. In fact the such a condition is necessary to guarantee the invertibility of the main matrix $M$. In fact, M is invertible only and only if $\Lambda$ is invertible which is equivalent to the admissible regularisation. In this case, it is possible to directly relate the Newton direction with the gradient and hence formulate directly the Hessian:
\begin{equation}
-PM^{-1}P^{T}\bar{N}=\bar{G}\quad\text{i.e.}\quad H=-PM^{-1}P^{T}
\end{equation}

This previous formula is of the utmost importance for the signature of the Hessian. If $P$ was invertible (which is never the case), then a direct application of Sylvester inertia theorem would have yielded that the Hessian and $M^{-1}$ (hence $M$) have the same signature i.e. signs of the eigenvalues. Unfortunately, the peripheral matrix $P$ is not full rank and we have to delve deeper to circumvent this limitation. The matrix $M$ is invertible and its inverse could be expressed through Schur complement:
\begin{equation}
    M^{-1}=\left[\begin{array}{cc}
\mathcal{I} & 0\\
-\Lambda^{-1}\left\{ \mathcal{I}-\mathcal{N}_{p}\right\} ^{T} & \mathcal{I}
\end{array}\right]\left[\begin{array}{cc}
-\left\{ \mathcal{I}-\mathcal{N}_{p}\right\} ^{-T}\Lambda\left\{ \mathcal{I}-\mathcal{N}_{p}\right\} ^{-1} & 0\\
0 & \Lambda^{-1}
\end{array}\right]\left[\begin{array}{cc}
\mathcal{I} & -\left\{ \mathcal{I}-\mathcal{N}_{p}\right\} \Lambda^{-1}\\
0 & \mathcal{I}
\end{array}\right]
\end{equation}

Let's call the following matrix $K$:
\begin{equation}
K=\left[\begin{array}{cc}
\mathcal{I} & 0\\
-\Lambda^{-1}\left\{ \mathcal{I}-\mathcal{N}_{p}\right\} ^{T} & \mathcal{I}
\end{array}\right]
\end{equation}

The Hessian expression involves product of $P$ and $K$ matrices. The idea at this stage is to transfer the diagonal term of the matrix $P$ towards the center. For this end it is possible to express the peripheral matrix $P$ as follows:
\begin{equation}
P=\left[\begin{array}{cc}
\nabla F_{p} & 0\\
\bar{X} & B
\end{array}\right]=\left[\begin{array}{cc}
\mathcal{I} & 0\\
\bar{X}\nabla F_{p}^{-1} & \mathcal{I}
\end{array}\right]\left[\begin{array}{cc}
\nabla F_{p} & 0\\
0 & B
\end{array}\right]
\end{equation}

The previous manipulation was possible thanks to the strict monotony of the activation function. We repeat the same procedure with the matrix product $PK$ and we get:
\begin{equation}
PK=\left[\begin{array}{cc}
\mathcal{I} & 0\\
-\bar{X}\nabla F_{p}^{-1}B\Lambda^{-1}\left\{ \mathcal{I}-\mathcal{N}_{p}\right\} ^{T}\nabla F_{p}^{-1} & \mathcal{I}
\end{array}\right]\left[\begin{array}{cc}
\nabla F_{p} & 0\\
0 & B
\end{array}\right]
\end{equation}

Thank to the previous formula we are finally able to to get the following LDL by block formula for the Stochastic Hessian:
\begin{equation}
\begin{cases}
H=Q\left[\begin{array}{cc}
\nabla F_{p}\left\{ \mathcal{I}-\mathcal{N}_{p}\right\} ^{-T}\Lambda\left\{ \mathcal{I}-\mathcal{N}_{p}\right\} ^{-1}\nabla F_{p} & 0\\
0 & -B\Lambda^{-1}B^{T}
\end{array}\right]Q^{T}\\
\\
Q=\left[\begin{array}{cc}
\mathcal{I} & 0\\
-\bar{X}\nabla F_{p}^{-1}B\Lambda^{-1}\left\{ \mathcal{I}-\mathcal{N}_{p}\right\} ^{T}\nabla F_{p}^{-1} & \mathcal{I}
\end{array}\right]
\end{cases}
\end{equation}

Q is lower triangular and invertible, hence the spectrum of the Hessian is given by:
\begin{equation}
    \mathcal{E}\left(M\right)=\mathcal{E}\left(-B\Lambda^{-1}B^{T}\right)\cup\mathcal{E}\left(\nabla F_{p}\left\{ \mathcal{I}-\mathcal{N}_{p}\right\} ^{-T}\Lambda\left\{ \mathcal{I}-\mathcal{N}_{p}\right\} ^{-1}\nabla F_{p}\right)
\end{equation}

The $\mathcal{E}$ refers to the ensemble of eigenvalues. Using properties of the backward operator the spectrum is further simplified as:
\begin{equation}
    \mathcal{E}\left(M\right)=\mathcal{E}\left(-B\Lambda^{-1}B^{T}\right)\cup\mathcal{E}\left(\nabla F_{p}\Lambda\nabla F_{p}\right)
\end{equation}

The previous formula shows that no matter what regularization we pick, both convex and concave directions are bound to coexist. Two families of eigenvalues are present within the stochastic Hessian. In the next section, we use the spectrum knowledge to reduce the radius of the Hessian to enhance the generalization ability of our method. 

\subsection{The search of generalization: targeting the flat minima}
Association between learning generalization and the concept of flat minima could be traced back to~\cite{hochreiter1995simplifying} and it has been extensively explored in the work of~\cite{jiang2019fantastic} . More recently SAM (Sharpness Awareness Minimisation)~\cite{foret2020sharpness} showed prominent validation rates for various data classifications. Within the context of the MK solution, we would like to confer a similar ability to our method by favoring flat minima during learning. We look to determine criteria to define the regularisation terms without any guesswork. Flat minima are associated with low curvature of the cost function landscape i.e. small amplitudes of the Hessian eigenvalues. We showed in the previous section that the Hessian has two families of eigenvalues (positive and negatives). In this section, we consider the absolute value of the eigenvalues and our goal is to minmax the absolute spectrum of the Hessian and therefore to reduce its radius. The two families of eigenvalues of the Hessian are the following (absolute value):
\begin{equation}
\begin{cases}
\mathcal{E}\left(\nabla F_{p}\Lambda\nabla F_{p}\right)\equiv\left\{ \lambda_{k}\left[\nabla F\left(a_{p}(k)\right)\right]_{i}^{2}\right\} _{k=1..n}\\
\\
\mathcal{E}\left(B\Lambda^{-1}B^{T}\right)\equiv\left\{ \frac{\left\Vert \nabla F\left(a_{p}(k)\right)b_{p}(k)\right\Vert _{2}^{2}}{\lambda_{k-1}}\right\} _{k=2..n}
\end{cases}
\end{equation}

The index $i$ in the previous formula indicates that we consider each entry in the vector (element-wise) for the maximization $\nabla F\left(a_{p}(k)\right)$. The maximum of the eigenvalues could be expressed as follows:
\begin{equation}
\max\left\{ \mathcal{E}\left(\nabla F_{p}\Lambda\nabla F_{p}\right),\text{\ensuremath{\mathcal{E}\left(B\Lambda^{-1}B^{T}\right)}}\right\} =\max_{k}\left\{ \lambda_{k}\left\Vert \nabla F\left(a_{p}(k)\right)\right\Vert _{\infty}^{2},\frac{\left\Vert \nabla F\left(a_{p}(k)\right)b_{p}(k)\right\Vert _{2}^{2}}{\lambda_{k-1}}\right\} 
\end{equation}

It is interesting to notice that each minimax procedure could be carried separately for each layer.
\begin{equation}
    \max_{k}\left\{ \lambda_{k}\left\Vert \nabla F\left(a_{p}(k)\right)\right\Vert _{\infty}^{2},\frac{\left\Vert \nabla F\left(a_{p}(k)\right)b_{p}(k)\right\Vert _{2}^{2}}{\lambda_{k-1}}\right\} =\max_{k}\max\left\{ \lambda_{k}\left\Vert \nabla F\left(a_{p}(k)\right)\right\Vert _{\infty}^{2},\frac{\left\Vert \nabla F\left(a_{p}(k+1)\right)b_{p}(k+1)\right\Vert _{2}^{2}}{\lambda_{k}}\right\} 
\end{equation}

Furthermore, for a given layer $k$ the values of $\left\Vert \nabla F\left(a_{p}(k)\right)\right\Vert _{\infty}^{2}$ and $\left\Vert \nabla F\left(a_{p}(k+1)\right)b_{p}(k+1)\right\Vert _{2}^{2}$ are fixed, hence varying the regularization parameter $\lambda_k$ would describe two curves (one linear and one hyperbolic) as shown in Figure~\ref{fig:eigenvalues}:
\begin{figure}[H]
    \centering
    \includegraphics[width=75mm]{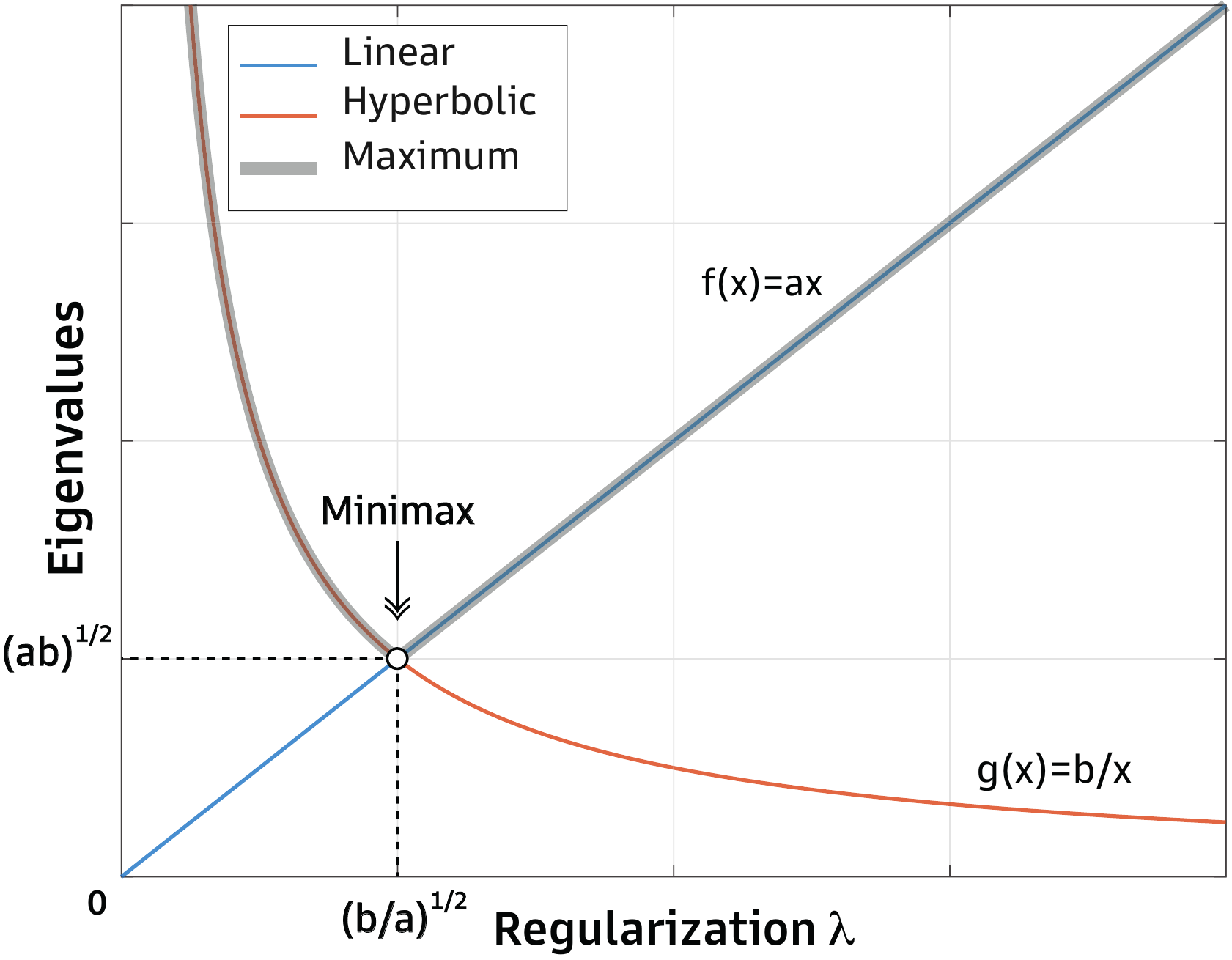}
    \caption{Curves of the two families of eigenvalues as a function of the regularization parameter for a hidden layer.}
    \label{fig:eigenvalues}
\end{figure}

With the previous solution we get a well defined criteria to select an optimal regularisation of all the hidden layers. Hence we have the following optimal regularization:
\begin{equation}
    \lambda_{k}=\frac{\left\Vert \nabla F\left(a_{p}(k+1)\right)b_{p}(k+1)\right\Vert _{2}}{\left\Vert \nabla F\left(a_{p}(k)\right)\right\Vert _{\infty}}\quad k=1..n-1
    \label{eq:regOpt}
\end{equation}

The determination of the $\left(\lambda_k\right)_{k=1..n-1}$ is a fixed point problem since the value of the adjoint $\left(b_p(k)\right)_k$ depends on the regularisation itself. In practice we will implement Equ.~(\ref{eq:regOpt}) using a non-regularized adjoint state vector. This approximation is asymptotically valid if convergence happens i.e. the backpropagated errors becomes small after some iterations. With this optimal definition the spectrum of the stochastic Hessian becomes:
\begin{equation}
   \rho\left(H\right)=\max\left\{ \left\Vert \nabla F\left(a_{p}(n)\right)\right\Vert _{\infty}^2,\left\Vert \nabla F\left(a_{p}(k+1)\right)b_{p}(k+1)\right\Vert _{2}\left\Vert \nabla F\left(a_{p}(k)\right)\right\Vert _{\infty}\right\}  
\end{equation}

The last layer is not concerned with this regularisation since the last diagonal term for the regularisation matrix $\Lambda$ is the unity. This creates a dissonance between the output layer and the rest of the network which is not favorable for our flat minima search. We address this last issue in the next section.

\subsection{The curious case of the output layer: towards an optimal step length}
The dissonance between the output layer and the regularization of the rest of the network is problematic since it would stop the Hessian from having a decreasing radius along the converging learning process. This anomaly is due to the way we introduced regularization with the following augmented cost function:
\begin{equation*}
    J=\frac{1}{2}\left\Vert d_{p}-x_{p}(n)\right\Vert _{2}^{2}+\sum_{k=1..n-1}\frac{\lambda_{k}}{2}\left\Vert x_{p}(k)\right\Vert _{2}^{2}
\end{equation*}

To circumvent the dissonance, we introduce the following fully regularized cost function:
\begin{equation}
    J=\frac{1}{2}\lambda_{n}\left\Vert d_{p}-x_{p}(n)\right\Vert _{2}^{2}+\sum_{k=1..n-1}\frac{\lambda_{k}}{2}\left\Vert x_{p}(k)\right\Vert _{2}^{2}
\end{equation}

Such a regularization of the last layer is akin to step length selection. Moreover, the two families of eigenvalues from the previous section do not constrain the new regularization parameter $\lambda_n$. We are therefore free to chose any value we deem suitable. We chose in particular the following value which corresponds to adding a white layer after the output:
\begin{equation}
    \lambda_{n}=\frac{\left\Vert b_{p}(n)\right\Vert _{2}}{\left\Vert \nabla F\left(a_{p}(n)\right)\right\Vert _{\infty}}
\end{equation}

This last value of regularisation yields a Hessian radius that would decrease along with convergence.
 \begin{equation}
\rho\left(H\right)=\max_{k=1..n-1}\left\{ \left\Vert \nabla F\left(a_{p}(k+1)\right)b_{p}(k+1)\right\Vert _{2}\left\Vert \nabla F\left(a_{p}(k)\right)\right\Vert _{\infty},\left\Vert b_{p}(n)\right\Vert _{2}\left\Vert \nabla F\left(a_{p}(n)\right)\right\Vert _{\infty}\right\} 
 \end{equation}
 To the best of our knowledge most of the activation functions used in Deep Learning have derivatives smaller than one, as such we can assert the following (large) inequality for the MK method:
 \begin{equation}
\rho\left(H\right)	\leq\max_{k=1..n}\left\{ \left\Vert b_{p}(k)\right\Vert _{2}\right\}  
 \end{equation}

With this definition of the output layer regularization/step we finally have all the elements for second-order stochastic Deep Learning, which is Cost-effective, handles Convexity issues, and is tuned to be Capable of generalization. We name this method the MK solution (short for Mehouachi and Kasmi).

\section{Applications}
\label{sec:Testcases} 
To validate the MK method for stochastic Deep Learning, two datasets are tested hereafter: MNIST and Fashion-MNIST, which are a widely referenced datasets in Machine Learning. We compare our MK method with Stochastic Gradient Descent (SGD) without any finetuning. MNIST and F-MNIST share the same image and characteristics. Both datasets are composed of a training set of 60,000 images and a test set of 10,000 images. Results are proposed and discussed in this section. It is worth reiterating that no finetuning has been performed to achieve better accuracy/validation and that the step has been maintained constant.

\subsection{MNIST dataset}
MNIST~\cite{LecunMNIST} database of handwritten digits was made available by Lecun et al. and is part of a larger data set available at NIST. It is composed of a training set of 60,000 hand-written digits images and a test set of 10,000 hand-written digits images. 

\begin{figure}[H]
    \centering
    \includegraphics[width=130mm]{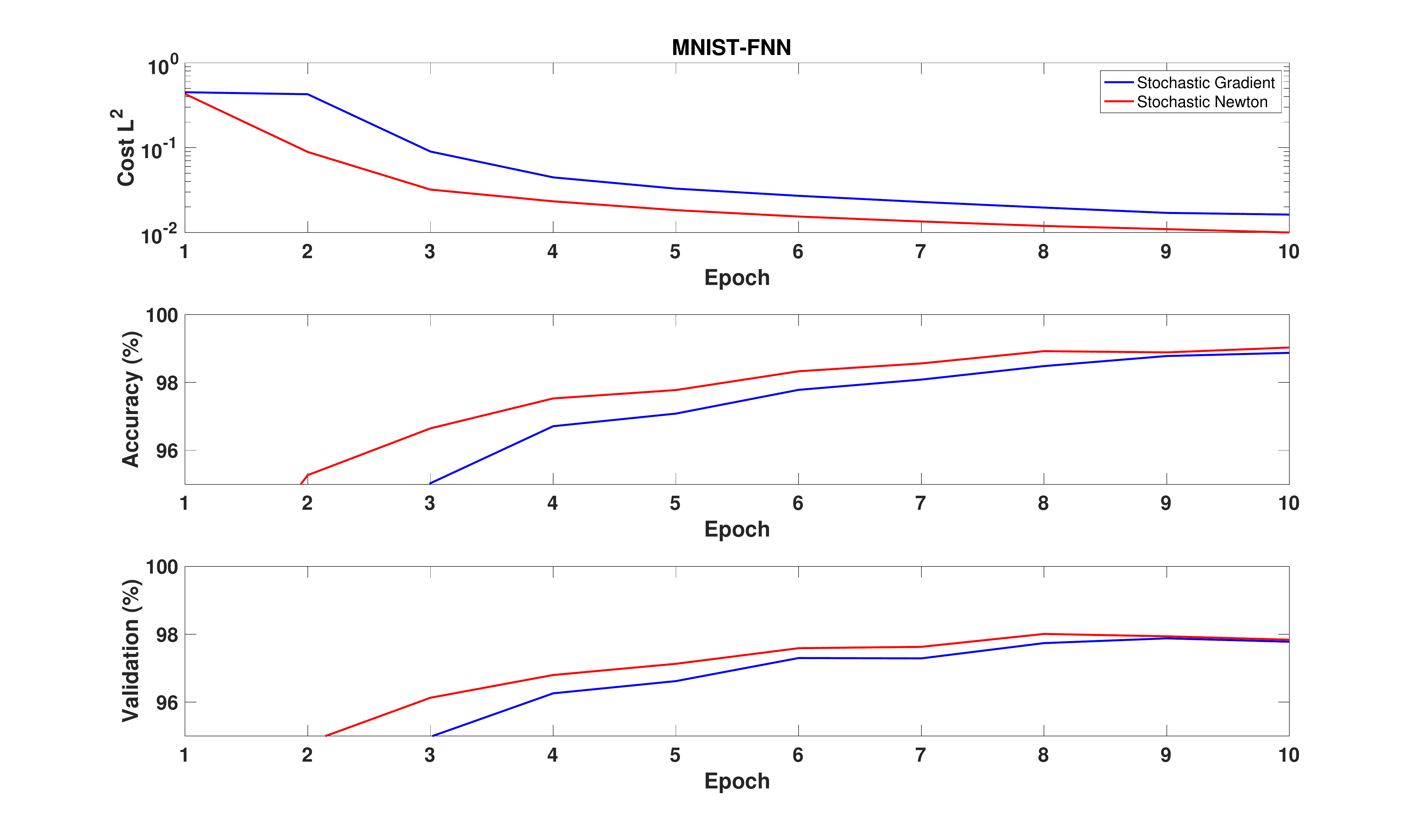}
    \caption{Cost function, Accuracy and Validation results for SGD and MK methods obtained for MNIST. The network has 3 hidden layers and the sizes are \{784 512 256 32 10\}.}
    \label{fig:mnist}
\end{figure}

\subsection{Fashion-MNIST dataset}
The Fashion-MNIST~\cite{xiao2017online} is a dataset composed of images of Zalando's articles. The Fashion-MNIST serves as a direct drop-in replacement for the original MNIST dataset and was made available for benchmarking Machine Learning algorithms.
\begin{figure}[H]
    \centering
    \includegraphics[width=115mm]{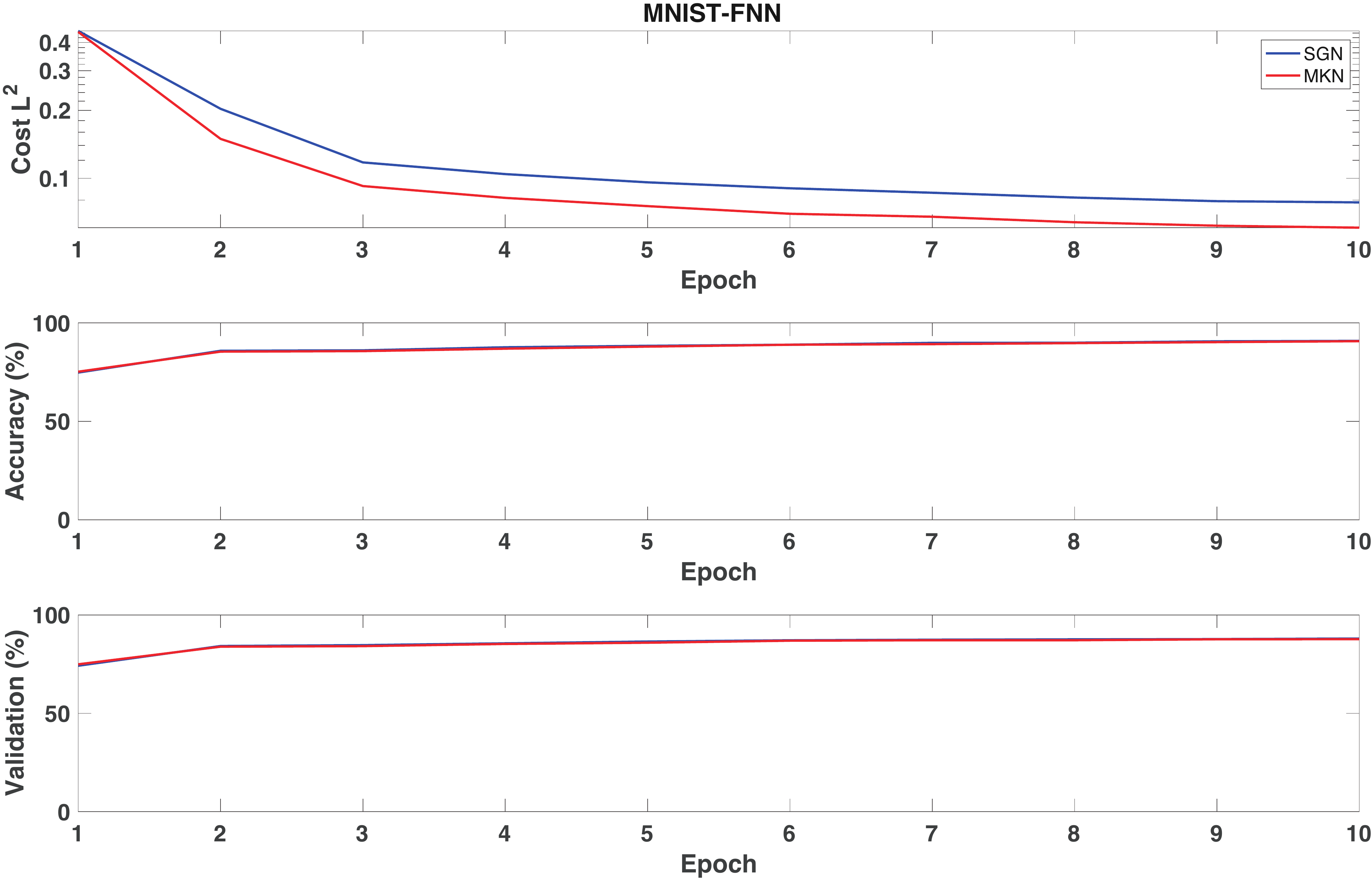}
    \caption{Cost function, Accuracy and Validation results for SGD and MK methods obtained for Fashion-MNIST. The network has 3 hidden layers and the sizes are \{784 512 256 32 10\}.}
    \label{fig:fmnist}
\end{figure}

\subsection{Advantages of the MK solution}
The different architecture of Neural Networks, e.g. number of hidden layers, number of neurons per layer, have been tested to investigate the advantage of the second-order approach. Interestingly, for so-called shallow architecture, the performances of the SGD and the stochastic Newton (MK) are quite similar. Nevertheless, when increasing the number of hidden layers, in the case of the deep neural network, we observed that the stochastic Newton (MK) performs better: faster convergence within a small number of epochs as shown Figs.~\ref{fig:mnist} and~\ref{fig:fmnist}. 

\section{Conclusion and Future Work}
\label{sec:conclusion}
In this paper, an exact stochastic second-order solution has been developed. The Sifrian tetrad system has been derived to provide a suitable foundation to investigate the advantages of the second order optimization for Deep Learning. The MK (Mehouachi-Kasmi) direction has been derived to deal with saddles and a robust spectral adjustment of the Hessian has been proposed to enhance convergence toward flat minima. Two applications, with a similar type of data sets namely MNIST and Fashion-MNIST, have been carried to test the performance of the MK method for second order Deep Learning.
This paper is a prelude to a full second order solution for Deep Learning that would handle the multi-pattern (batch) case, different forms of regularization and would open the frontiers for a higher order Deep Learning (Halley Deep Learning).

\section*{Acknowledgements}
The authors would like to thank Dr. Najwa Aaraj, Chief Researcher at Technology Innovation Institute, and Dr. Merouane Debbah, Prof. at Centralesupelec for their valuable feedback and suggestions. Moreover, the authors are grateful to Technology Innovation Institute (https://www.tii.ae/) for providing the necessary environment to explore avant-garde concepts in Artificial Intelligence.

\section*{Disclaimer}
The study presented in this paper is a work in progress at the Directed Energy Research Centre of the Technology Innovation Institute of Abu Dhabi. The present manuscript will be subject to modifications and improvements as it is not yet peer-reviewed. The authors welcome comments and suggestions from AI researchers. Please do not hesitate to contact the authors for more details. Further results will be made available soon.

\bibliographystyle{unsrt}  
\bibliography{references} 
\end{document}